\def\year{2020}\relax
\documentclass[letterpaper]{article} % DO NOT CHANGE THIS
\usepackage{aaai20}  % DO NOT CHANGE THIS
\usepackage{times}  % DO NOT CHANGE THIS
\usepackage{helvet} % DO NOT CHANGE THIS
\usepackage{courier}  % DO NOT CHANGE THIS
\usepackage[hyphens]{url}  % DO NOT CHANGE THIS
\usepackage{graphicx} % DO NOT CHANGE THIS
\usepackage{graphicx}  % DO NOT CHANGE THIS
\usepackage{multirow}
\usepackage{todonotes}
\usepackage{amsmath}
\usepackage{subcaption}
\newcommand{\citet}[1]{\citeauthor{#1} \shortcite{#1}}
\title{Attention-Informed Mixed-Language Training for \\ Zero-shot Cross-lingual Task-oriented Dialogue Systems}
\author{Zihan Liu\thanks{The authors contributed equally to this work.}, Genta Indra Winata$^*$, Zhaojiang Lin, Peng Xu, Pascale Fung \\ % All authors must be in the same font size and format. Use \Large and \textbf to achieve this result when breaking a line
Center for Artificial Intelligence Research (CAiRE)\\ 
The Hong Kong University of Science and Technology
\\ %If you have multiple authors and multiple affiliations
\{zliucr, giwinata, zlinao, pxuab\}@connect.ust.hk, pascale@ece.ust.hk % email address must be in roman text type, not monospace or sans serif
}
 
\begin{document}

\maketitle

\begin{abstract}
Recently, data-driven task-oriented dialogue systems have achieved promising performance in English. However, developing dialogue systems that support low-resource languages remains a long-standing challenge due to the absence of high-quality data. In order to circumvent the expensive and time-consuming data collection, we introduce Attention-Informed \textbf{M}ixed-\textbf{L}anguage \textbf{T}raining (\textbf{MLT}), a novel zero-shot adaptation method for cross-lingual task-oriented dialogue systems. It leverages very few task-related parallel word pairs to generate code-switching sentences for learning the inter-lingual semantics across languages. Instead of manually selecting the word pairs, we propose to extract source words based on the scores computed by the attention layer of a trained English task-related model and then generate word pairs using existing bilingual dictionaries. Furthermore, intensive experiments with different cross-lingual embeddings demonstrate the effectiveness of our approach. Finally, with very few word pairs, our model achieves significant zero-shot adaptation performance improvements in both cross-lingual \textit{dialogue state tracking} and \textit{natural language understanding} (i.e., intent detection and slot filling) tasks compared to the current state-of-the-art approaches, which utilize a much larger amount of bilingual data.
\end{abstract}

% Introduction
% 1. The requirement of task-oriented dialogue systems. Achievement in task-oriented dialogue system
% 2. The requirement of dialogue systems in other languages and the current problem (lack enough data).
% 3. How to cope with this problem.
% 4. Describe the problems of current dialogue systems
% 5. Introduce our method, leverage mixed-language training (only require a little amount of word pairs), and why our method can work. And contributions of our work.

% 1. What are the issues?
\section{Introduction}
Over the past few years, the demand of task-oriented dialogue systems has increased rapidly across the world, following their promising performance on English systems~\cite{zhong2018global,wu2019transferable}.
However, most dialogue systems are unable to support numerous low-resource languages due to the scarcity of high-quality data, which will eventually create a massive gap between the performance of low-resource language systems (e.g., Thai) and high-resource systems (e.g., English).
A common straightforward strategy to address this problem is to collect more data and train each monolingual dialogue system separately, but it is costly and resource-intensive to collect new data on every single language.

% 2. How to cope with this problem
Zero-shot adaptation is an effective approach to circumvent the data collection process when there is no training data available by transferring the learned knowledge from a high-resource source language to low-resource target languages. Currently, a few studies have been performed on the \textit{zero-shot learning} in task-oriented dialogue systems~\cite{chen2018xl,schuster2019cross}.
However, there are two problems that exist in this research: 
(1) the existing methods require a sufficient parallel corpus, which is not ideal for training models on rare languages where bilingual resources are minimal, and (2) the imperfect alignments of cross-lingual embeddings such as MUSE~\cite{conneau2017word} as well as the enormous cross-lingual models XLM~\cite{lample2019cross}, and Multilingual BERT~\cite{devlin2019bert} limit the cross-lingual zero-shot transferability.

\begin{figure*}[ht!]
    \centering
    \includegraphics[scale=0.82]{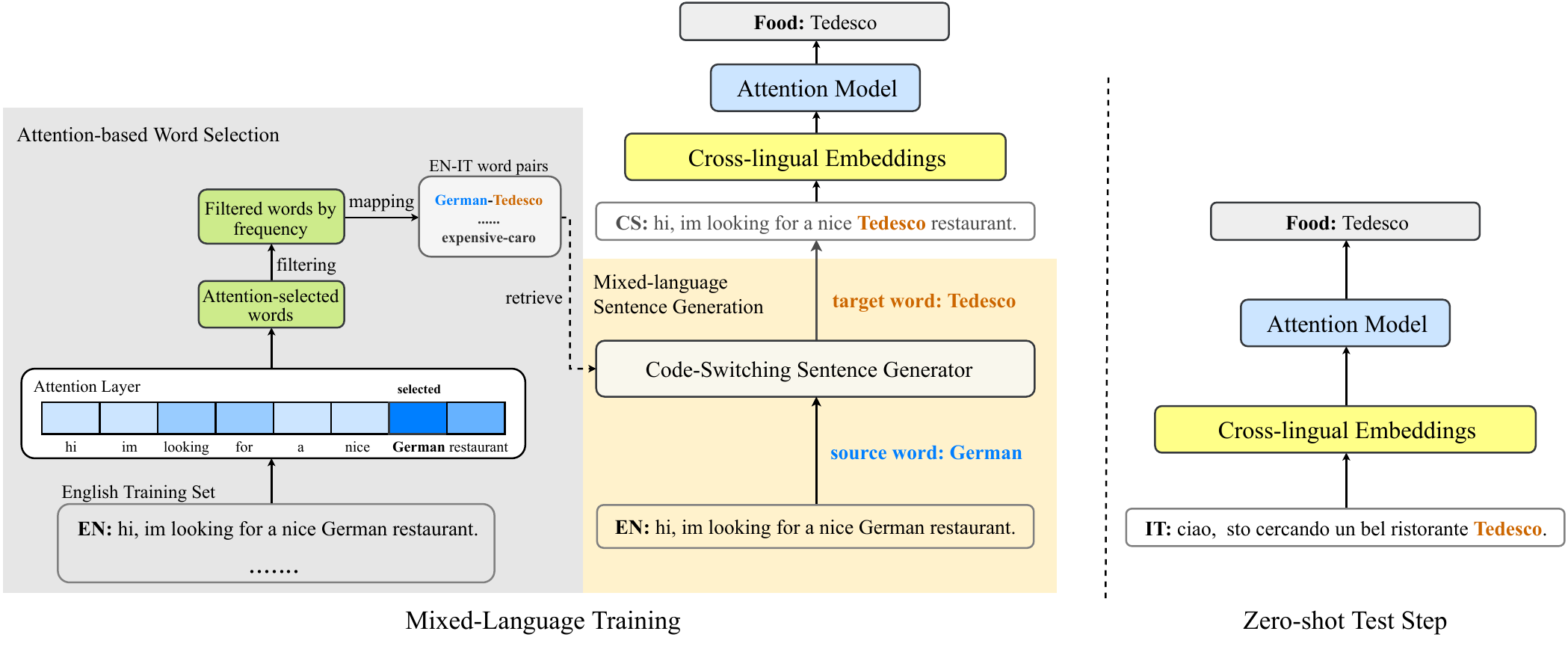}
    \caption{Illustration of the mixed-language training (MLT) approach and zero-shot transfer. \textbf{EN} denotes an English text, \textbf{IT} denotes an Italian text, and \textbf{CS} denotes a code-switching text (i.e., a mixed-language sentence). In the training step, code-switching sentence generator will replace the task-related word with its corresponding translation in the target language to generate code-switching sentences. In the zero-shot transfer step, we leverage cross-lingual word embeddings and directly adapt the trained attention model to the target language.}
    \label{fig:MLT}
\end{figure*}

% 5. Introduce our method, leverage mixed-language training (only require a little amount of word pairs), and why our method can work. And contributions of our work.
To address these problems, we propose the \textbf{attention-informed mixed-language training} \textbf{(MLT)}, a new framework that leverages extremely small number of bilingual word pairs to build zero-shot cross-lingual task-oriented dialogue systems. The word pairs are created by choosing words from the English training data using attention scores from a trained English model. Then we pair these English words with target words using existing bilingual dictionaries, and use the target words to replace keywords in the training data and build code-switching sentences.\footnote{``code-switching" is interchangeable with ``mixed-language".} The intuition behind training with code-switching sentences is to help the model to identify selected important keywords as well as their semantically similar keywords in the target language. In addition, we incorporate the MUSE, RCSLS~\cite{joulin2018loss}, and cross-lingual language models XLM and Multilingual BERT for generating cross-lingual embeddings.

During the training phase, our model learns to capture important keywords in code-switching sentences mixed with source and target language words. We conjuncture that learning with task-related keywords of the target language helps the model to capture other task-related words that have similar semantics, for example, synonyms or words in the same category such as days of the week ``Domingo'' (Sunday) and ``Lunes'' (Monday). During the zero-shot testing phase, the inter-lingual understanding learned by the model alleviates the main issue of the imperfect alignment of cross-lingual embeddings. The experimental results on unseen languages show that MLT outperforms existing baselines with significant margins in both dialogue state tracking and natural language understanding tasks on all languages using many fewer resources. This proves that our approach is effective for application to low-resource languages when there is only limited parallel data available.\footnote{The code is available at: https://github.com/zliucr/mixed-language-training}

Contributions in our work are summarized as follows:
\begin{itemize}
    \item We investigate the extremely low bilingual resources setting for zero-shot cross-lingual task-oriented dialogue systems.
    \item Our approach achieves state-of-the-art zero-shot cross-lingual performance in both \textit{dialogue state tracking} and \textit{natural language understanding} of task-oriented dialogue systems using many fewer bilingual resources.
    \item We study the performance of current cross-lingual pre-trained language models (namely Multilingual BERT and XLM) on zero-shot cross-lingual dialogue systems, and conduct quantitative analyses while adapting them to cross-lingual dialogue systems.
\end{itemize}

\section{Related Work}
\subsection{Task-oriented Dialogue Systems}
Dialogue state tracking (DST) and natural language understanding (NLU) are the key components for understanding user inputs and building dialogue systems.

\subsubsection{Dialogue State Tracking}
\citet{mrkvsic2017neural} proposed to utilize pre-trained word vectors by composing them into a distributed representation of user utterances and to resolve morphological ambiguity. \citet{zhong2018global} successfully improved rare slot values tracking through slot-specific local modules. 

\subsubsection{Natural Language Understanding}
\citet{liu2016attention} leveraged an attention mechanism to learn where to pay attention in the input sequences for joint intent detection and the slot filling task. \citet{goo2018slot} introduced slot-gated models to learn the relationship between intent and slot attention vectors and better captured the semantics of user utterances and queries.

\subsubsection{Multilingual Task-oriented Dialogue Systems}
A number of multilingual task-oriented dialogue systems datasets have been published lately~\cite{mrkvsic2017semantic,schuster2019cross}, enabling evaluation of the approaches for cross-lingual dialogue systems. \citet{mrkvsic2017semantic} annotated two languages (namely German and Italian) for the dialogue state tracking dataset \textbf{WOZ 2.0}~\cite{mrkvsic2017neural} and trained a unified framework to cope with multiple languages. Meanwhile, \citet{schuster2019cross} introduced a multilingual NLU dataset and highlighted the need for more sophisticated cross-lingual methods.

\begin{figure*}[!ht]
\centering
\includegraphics[scale=0.72]{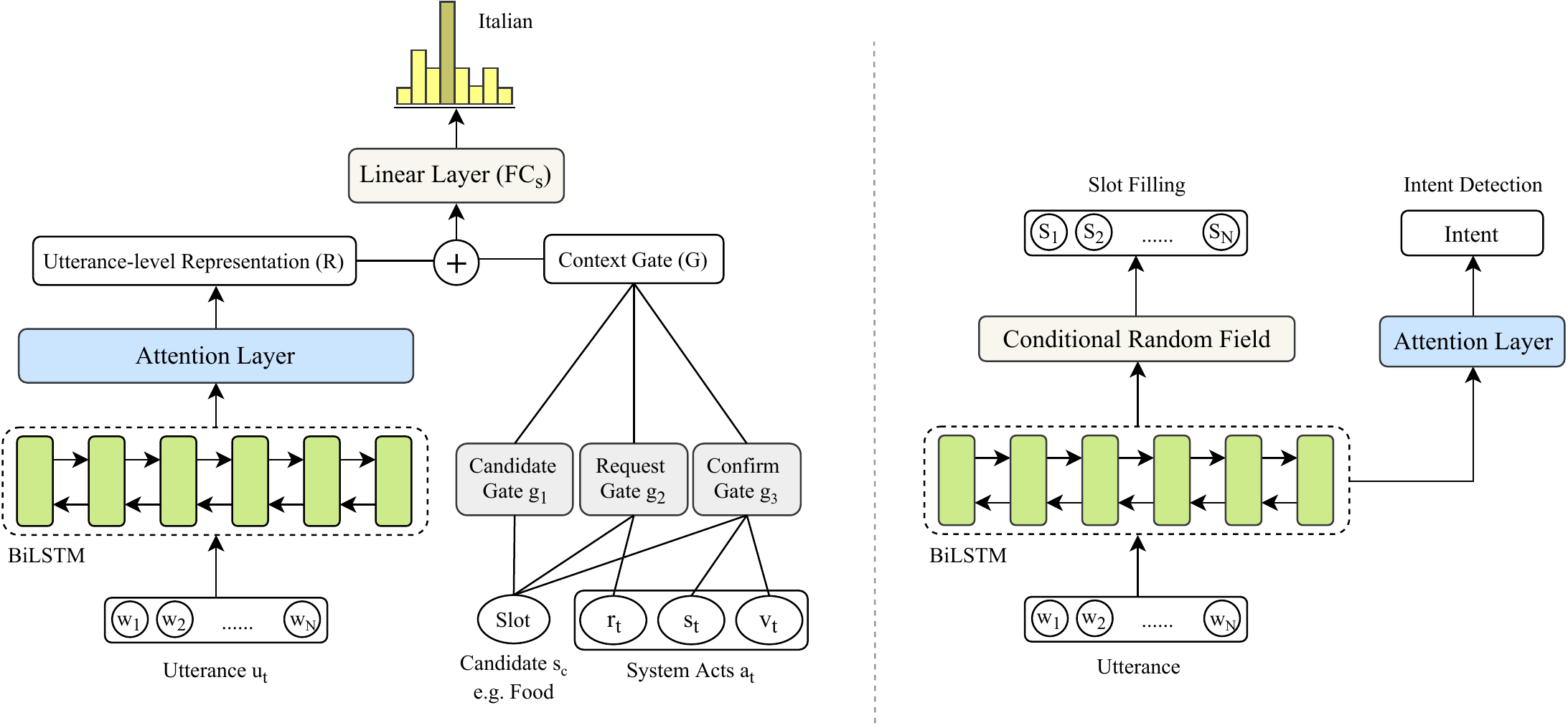}
\caption{Dialogue State Tracking Model \textbf{(left)} and Natural Language Understanding Model \textbf{(right)}. For each model, we apply an attention layer to learn important task-related words.}
\label{fig:models}
\end{figure*}

\subsection{Cross-lingual Transfer Learning}
Cross-lingual transfer learning, which aims to discover the underlying connections between the source and target language, has become a popular topic recently. \citet{conneau2017word} proposed to use zero supervision signals to conduct cross-lingual word embedding mapping and achieved promising results.
\citet{devlin2019bert,lample2019cross} leveraged large monolingual and bilingual corpus to align cross-lingual sentence-level representations and achieved the state-of-the-art performance in many cross-lingual tasks.
Recently, studies have applied cross-lingual transfer algorithms to natural language processing tasks, such as \textit{named entity recognition} (NER) \cite{ni2017weakly}, \textit{entity linking} \cite{pan2017cross}, \textit{POS tagging} \cite{kim2017cross,zhang2016ten}, and \textit{dialogue systems} \cite{chen2018xl,upadhyay2018almost,liu2019zero}.
Nevertheless, to the best of our knowledge, only a few studies have focused on task-oriented dialogue systems, and none of them investigated the extremely low bilingual resources scenario.

\section{Mixed-Language Training}
% 1. Task-related words selection
% 2. Mixed-language sentences generator
As shown in Figure \ref{fig:MLT}, in the mixed-language training step, our model is trained using code-switching sentences generated from source language sentences by replacing the selected source words with their translations. In the zero-shot test step, our model directly transfers into the unseen target language.

\subsection{Attention-based Selection}
Intuitively, the attention layer in a trained model can focus on the keywords that are related to the task. As shown in Figure~\ref{fig:MLT}, we propose to utilize the scores computed from the attention layer of a trained model on source language (English) data to select keywords for completing the task. 
Concretely, we first collect source words by taking the top-1 attention score for each source utterance since the source words with the highest attention score are the most important for the given task. However, some noisy words (unimportant words) might still exist in the collection. Hence, we first count the times that the words are selected and filter the words that are seldom selected, and then we choose the top-$n$ most frequent words in the training set as our final word candidates and pair them using an existing bilingual dictionary. We denote the selected $n$ word pairs as a key-value dictionary $D =  ((x_1;y_1),\dots,(x_n;y_n))$, where $x$ and $y$ represent the source and target language, respectively.

% 2. Mixed-language sentences generator
% \subsection{Code-Switching Sentence Generator}
\subsection{Training and Adaptation}

Given a source language sentence $\mathbf{w}=[w_1,w_2,\dots,w_N]$, we replace the words in $\mathbf{w}$ with their corresponding target words if they are present in $D$ to generate a code-switching sentence $\mathbf{w}_{cs}$.
As illustrated in Figure \ref{fig:MLT}, we use cross-lingual word embeddings for source and target language words. 
\begin{equation}
    \mathbf{w}_{cs} = \textnormal{CS}_{gen}(\mathbf{w}),
\end{equation}
\begin{equation}
    out = \textnormal{AttnModel}(E(\mathbf{w}_{cs})),
\end{equation}
where CS$_{gen}$ represents the code-switching sentence generator in Figure \ref{fig:MLT}, \textit{AttnModel} represents the attention model, and $E$ denotes cross-lingual word embeddings. We specifically use cross-lingual word embeddings from MUSE~\cite{conneau2017word} and RCSLS~\cite{joulin2018loss}, aligned representations of source and target languages to transfer the learned knowledge from the source language to the target language. By applying mixed-language training, our model can cope with the problem of imperfect alignment of cross-lingual word embeddings. In the zero-shot test step, the attention layer is still able to focus on the same or semantically similar target language keywords, as it does in the mixed-language training step, which improves the robustness of cross-lingual transferability.

\section{Cross-lingual Dialogue Systems}
In this section, we focus on applying our mixed-language training approach to cross-lingual task-oriented dialogue systems. We design model architectures for dialogue state tracking and natural language understanding (i.e., intent detection and slot filling) as follows.

\subsection{Dialogue State Tracking}
% 1. Describe our model, use formula
% 2. Analyze the strength of our model
Our dialogue state tracking (DST) model, illustrated in Figure \ref{fig:models}, is modified from \citet{chen2018xl}. We model DST into a classification problem based on three inputs: (i) the user utterance $ u_t $, (ii) the slot candidate $ s_c $, and (iii) the system dialogue acts $ a_t = (r_t, s_t, v_t) $\footnote{$r_t$ represents the system request, and $s_t$ and $v_t$ represent the system confirmation. For example, when the system requests more information by asking ``Do you have an area preference?'', then \textbf{$r_t$ = ``area''}, or when the system confirms by saying ``The Vietnamese food is in the cheap price range,'' then $s_t$ = \textbf{``price range''} and $v_t$ = \textbf{``cheap''}.}, where we use subscript $t$ to denote each dialogue turn. In short, our model can be decomposed into the following three components:

\subsubsection{Utterance Encoder}
We use a bi-directional LSTM (BiLSTM) to encode the user utterance $u_t=[w_1, w_2 \dots w_N]$ and an attention mechanism~\cite{felbo2017using} on top of the BiLSTM to generate an utterance representation $R$, where $w_i$ is the word vector of the $i$-th token and $N$ is the length of the utterance. We formalize the utterance encoder as:
\begin{equation}
    [h_1, h_2 \dots h_N] = \textnormal{BiLSTM}([w_1, w_2 \dots  w_N]),
\end{equation}
\begin{equation} \label{eq5}
    e_i = h_i w_a, \hspace{2mm} 
    \alpha_i = \frac{exp(e_i)}{\sum_{j=1}^N exp(e_j)}, \hspace{2mm} R = \sum_{i=1}^N \alpha_i h_i,
\end{equation}
where $w_a$ is a trainable weight vector in the attention layer, and $\alpha_i$ is the attention score of each token $i$.

\subsubsection{Context Gate}
Given a candidate slot $s_c$ and system acts $(r_t, s_t, v_t)$ as inputs, we compute the context gate $G$ by summing three individual gates: (i) the candidate gate ($g_1$), (ii) the request gate ($g_2$), and (iii) the confirm gate ($g_3$). The context gate is defined as follows:
\begin{equation}
    g_1 = E(s_c), \hspace{3mm} g_2 = \sigma(E(s_c) \odot W_1 E(r_t)),
\end{equation}
\begin{equation}
    g_3 = \sigma(E(s_c) \odot W_2 (E(s_t)+E(v_t))),
\end{equation}
\begin{equation}
    G = g_1 + g_2 + g_3,
\end{equation}
where $E$ denotes the word embedding look-up table, $\odot$ denotes a Hadamard product, $W_1$ and $W_2$ represent trainable parameter matrices, and $\sigma$ represents a sigmoid function.

\subsubsection{Slot Value Prediction}
Finally, we concatenate the utterance representation ($R$) and the context gate ($G$), which are then passed into a linear layer $FC_s$ and a softmax layer for prediction.

\subsection{Natural Language Understanding}
Our NLU model is illustrated in Figure \ref{fig:models} as a multi-task problem. We describe our model as follows:

\subsubsection{Slot Filling} 
We use a BiLSTM-CRF combining a BiLSTM with a conditional random field (CRF) sequence labeling model~\cite{lample2016neural} for slot prediction. We pass the hidden states of the BiLSTM through a softmax layer and then pass the resulting label probability vectors through the CRF layer for computing final predictions. 

\subsubsection{Intent Prediction} 
We place an attention layer over the hidden states of the BiLSTM and predict the intent for the user utterance through a softmax projection layer. The attention layer is similar to the one in the dialogue state tracking shown in equation (\ref{eq5}). 
% need to explain the detail of word pairs we choose for NLU

\begin{figure}
    \centering
    \includegraphics[scale=0.67]{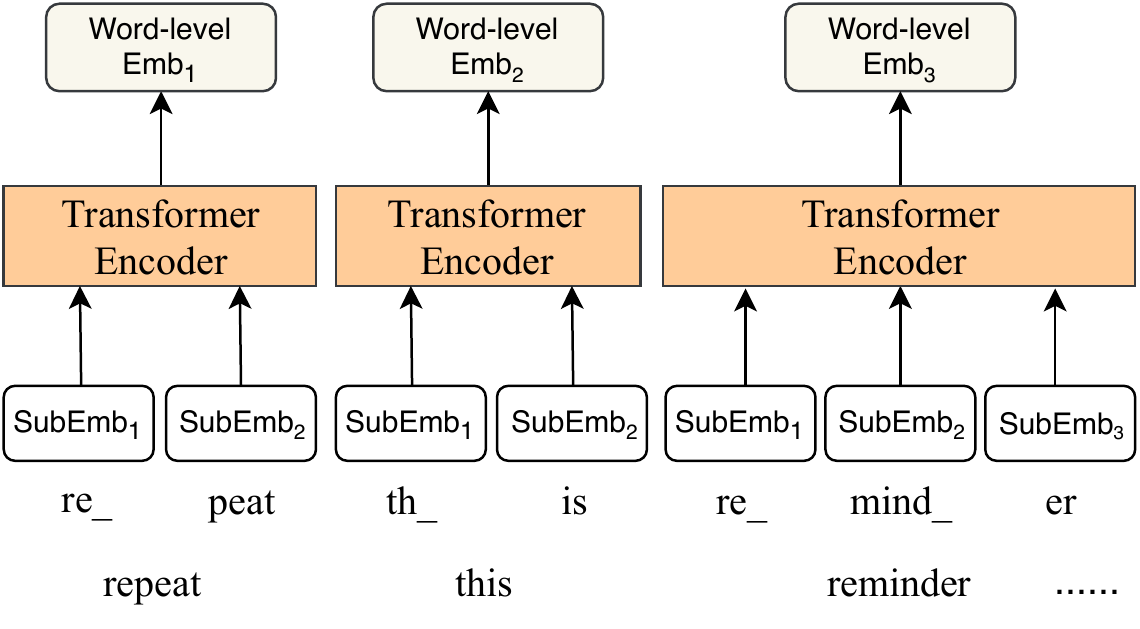}
    \caption{Illustration of how we leverage a \textit{transformer encoder} to incorporate subword embeddings into word-level representations. The parameters in the \textit{transformer encoder} are shared for all subword embeddings.}
    \label{fig:subword2word}
\end{figure}

\begin{table*}[!ht]
\centering
\resizebox{0.82\textwidth}{!}{
\begin{tabular}{l|ccc|ccc|ccc}
\hline
\multicolumn{1}{c|}{} & \multicolumn{9}{c}{\textbf{German}} \\ \hline
\multicolumn{1}{c|}{\multirow{2}{*}{Model}} & \multicolumn{3}{c|}{\textbf{slot acc.}} & \multicolumn{3}{c|}{\textbf{joint goal acc.}} & \multicolumn{3}{c}{\textbf{request acc.}} \\ \cline{2-10} 
\multicolumn{1}{c|}{} & \textbf{BASE} & \textbf{MLT}$_O$ & \textbf{MLT}$_A$ & \textbf{BASE} & \textbf{MLT}$_O$ & \textbf{MLT}$_A$ & \textbf{BASE} & \textbf{MLT}$_O$ & \textbf{MLT}$_A$ \\ \hline
MUSE & 60.69 & 68.58 & \textbf{71.38} & 21.57 & 30.61 & \textbf{36.51} & 74.22 & 80.11 & \textbf{82.99} \\ \hline
XLM (MLM)$^\ast$ & 52.21 & 66.26 & \textbf{68.25} & 14.09 & 29.45 & \textbf{31.29} & 75.15 & 78.48 & \textbf{80.22} \\
\quad+ Transformer & 53.81 & 65.81 & \textbf{68.55} & 13.97 & 30.87 & \textbf{32.98} & 76.83 & 78.95 & \textbf{81.34} \\ \hline
XLM (MLM+TLM)$^\ast$ & 58.04 & 65.39 & \textbf{66.25} & 16.34 & 29.22 & \textbf{29.83} & 75.73 & 78.86 & \textbf{79.12} \\
\quad+ Transformer & 56.52 & 66.81 & \textbf{68.88} & 16.59 & 31.76 & \textbf{33.12} & 78.56 & 81.59 & \textbf{82.96} \\ \hline
Multi. BERT$^\ast$ & 57.61 & 67.49 & \textbf{69.48} & 14.95 & 30.69 & \textbf{32.23} & 75.31 & 83.66 & \textbf{86.27} \\ 
\quad+ Transformer & 57.43 & 68.33 & \textbf{70.77} &  15.67 & 31.28 & \textbf{34.36} & 78.59 & 84.37 & \textbf{86.97} \\ \hline \hline
\textit{Ontology Matching$^\dagger$} & \multicolumn{3}{c|}{\textit{24}} & \multicolumn{3}{c|}{\textit{-}} &  \multicolumn{3}{c}{\textit{21}} \\ 
\textit{Translate Train$^\dagger$} & \multicolumn{3}{c|}{\textit{41}} & \multicolumn{3}{c|}{\textit{-}} &  \multicolumn{3}{c}{\textit{42}} \\ \hline
\textit{Bilingual Dictionary$^\ddagger$} & \multicolumn{3}{c|}{\textit{51.74}} & \multicolumn{3}{c|}{\textit{28.07}} &  \multicolumn{3}{c}{\textit{72.54}} \\
\textit{Bilingual Corpus$^\ddagger$ } & \multicolumn{3}{c|}{\textit{55}} & \multicolumn{3}{c|}{\textit{30.84}} &  \multicolumn{3}{c}{\textit{68.32}} \\ \hline
\textit{Supervised Training} & \multicolumn{3}{c|}{\textit{85.78}} & \multicolumn{3}{c|}{\textit{78.89}} &  \multicolumn{3}{c}{\textit{84.02}}\\ \hline \hline
\multicolumn{1}{c|}{} & \multicolumn{9}{c}{\textbf{Italian}} \\ \hline
\multicolumn{1}{c|}{\multirow{2}{*}{Model}} & \multicolumn{3}{c|}{\textbf{slot acc.}} & \multicolumn{3}{c|}{\textbf{joint goal acc.}} & \multicolumn{3}{c}{\textbf{request acc.}} \\ \cline{2-10} 
\multicolumn{1}{c|}{} & \textbf{BASE} & \textbf{MLT}$_O$ & \textbf{MLT}$_A$ & \textbf{BASE} & \textbf{MLT}$_O$ & \textbf{MLT}$_A$ & \textbf{BASE} & \textbf{MLT}$_O$ & \textbf{MLT}$_A$ \\ \hline
MUSE & 60.59 & 73.55 & \textbf{76.88} & 20.66 & 36.88 & \textbf{39.35} & 79.09 & 82.24 &\textbf{84.23} \\ \hline
Multi. BERT$^\ast$ & 53.34 & 65.49 & \textbf{69.48} & 12.88 & 26.45 & \textbf{31.41} & 76.12 & 84.58 & \textbf{85.18} \\ 
\quad+ Transformer & 54.56 & 66.87 & \textbf{71.45} & 12.63 & 28.59 & \textbf{33.35} & 77.34 & 82.93 & \textbf{84.96} \\ \hline \hline
\textit{Ontology Matching$^\dagger$} & \multicolumn{3}{c|}{\textit{23}} & \multicolumn{3}{c|}{\textit{-}} & \multicolumn{3}{c}{\textit{21}} \\ 
\textit{Translate Train$^\dagger$} & \multicolumn{3}{c|}{\textit{48}} & \multicolumn{3}{c|}{\textit{-}} & \multicolumn{3}{c}{\textit{51}} \\ \hline
\textit{Bilingual Dictionary$^\ddagger$} &  \multicolumn{3}{c|}{\textit{73}} & \multicolumn{3}{c|}{\textit{39.01}} & \multicolumn{3}{c}{\textit{77.09}} \\
\textit{Bilingual Corpus$^\ddagger$ } & \multicolumn{3}{c|}{\textit{72}} & \multicolumn{3}{c|}{\textit{41.23}} & \multicolumn{3}{c}{\textit{81.23}} \\ \hline
\textit{Supervised Training} &  \multicolumn{3}{c|}{\textit{88.92}} & \multicolumn{3}{c|}{\textit{80.22}} & \multicolumn{3}{c}{\textit{91.05}} \\ \hline
\end{tabular}
}
\caption{Zero-shot results for the target languages on \textbf{Multilingual WOZ 2.0}. \textbf{MLT}$_A$ denotes our approach (attention-informed MLT), which utilizes the same number of word pairs as \textbf{MLT}$_O$ (90 word pairs). $^\ddagger$ denotes the results of XL-NBT. Note that, we realize that the goal accuracy in \citet{chen2018xl} is calculated as slot accuracy in our paper, so we rerun the models using the provided code (https://github.com/wenhuchen/Cross-Lingual-NBT) to calculate joint goal accuracy. $^\dagger$ denotes the results from \citet{chen2018xl}. Instead of using the \textit{transformer encoder}, we sum the subword embeddings based on the word boundaries to get word-level representations. Due to the absence of the Italian language in the XLM models, we cannot report the results.}
\label{table:dst}
\end{table*}

\section{Cross-lingual Language Model}
We investigate the effectiveness of current powerful cross-lingual pre-trained language models XLM and Multilingual BERT, and deploy MLT into them for the zero-shot cross-lingual DST and NLU tasks. \citet{lample2019cross} proposed cross-lingual language model pre-training (XLM) and two objective functions \textit{masked language modeling} (MLM) and \textit{translation language modeling} (TLM). The MLM leveraged a monolingual corpus, the TLM utilized a bilingual corpus, and MLM+TLM incorporated both MLM and TLM. Pre-trained XLM models on 15 languages are publicly available.\footnote{https://github.com/facebookresearch/XLM} Multilingual BERT is trained on the monolingual corpora of 104 languages, and the model is also publicly available.\footnote{https://github.com/google-research/bert/blob/master/multi- lingual.md}

In order to handle multiple languages and reduce the vocabulary size, both methods leverage subword units to tokenize each sentence.
However, the outputs of the DST and NLU tasks depend on the word-level information.
Hence, we propose to learn the mapping between the subword-level and word-level by adding a transformer encoder~\cite{dehghani2018universal} on top of subword units and learn to encode them into word-level embeddings, which we describe in Figure \ref{fig:subword2word}. After that, we leverage the same model structures as illustrated in Figure \ref{fig:models} for the DST and NLU tasks. 

\section{Experiments}
\subsection{Datasets}
\subsubsection{Dialogue State Tracking} Wizard of Oz (WOZ), a restaurant domain dataset, is used for training and evaluating dialogue state tracking models on English. It was enlarged into WOZ 2.0 by adding more dialogues, and recently, \citet{mrkvsic2017semantic} expanded WOZ 2.0 into Multilingual WOZ 2.0 by including two more languages (German and Italian). Multilingual WOZ 2.0 contains 1200 dialogues for each language, where 600 dialogues are used for training, 200 for validation, and 400 for testing. The corpus contains three goal-tracking slot types: food, price range and area, and a request slot type. The model has to track the value for each goal-tracking slot and request slot.

\subsubsection{Natural Language Understanding} Recently, a multilingual task-oriented natural language understanding dialogue dataset was proposed by~\citet{schuster2019cross}, which contains English, Spanish, and Thai across three domains (alarm, reminder, and weather). The corpus includes 12 intent types and 11 slot types, and the model has to detect the intent of the user utterance and conduct slot filling for each word of the utterance.

\begin{table*}[!ht]
\centering
\resizebox{\textwidth}{!}{
\begin{tabular}{l|ccc|ccc|ccc|ccc}
\hline
\multicolumn{1}{c|}{} & \multicolumn{6}{c|}{\textbf{Spanish}} & \multicolumn{6}{c}{\textbf{Thai}} \\ \hline
\multicolumn{1}{c|}{\multirow{2}{*}{Model}} & \multicolumn{3}{c|}{\textbf{Intent acc.}} & \multicolumn{3}{c|}{\textbf{Slot F1}} &  \multicolumn{3}{c|}{\textbf{Intent acc.}} & \multicolumn{3}{c}{\textbf{Slot F1}} \\ \cline{2-13}
\multicolumn{1}{c|}{} & \multicolumn{1}{l}{\textbf{BASE}} & \multicolumn{1}{l}{\textbf{MLT}$_H$} & \multicolumn{1}{l|}{\textbf{MLT}$_{A}$} & \multicolumn{1}{l}{\textbf{BASE}} & \multicolumn{1}{l}{\textbf{MLT}$_H$} & \multicolumn{1}{l|}{\textbf{MLT}$_A$} &  \multicolumn{1}{l}{\textbf{BASE}} & \multicolumn{1}{l}{\textbf{MLT}$_H$} & \multicolumn{1}{l|}{\textbf{MLT}$_A$} & \multicolumn{1}{l}{\textbf{BASE}} & \multicolumn{1}{l}{\textbf{MLT}$_H$} & \multicolumn{1}{l}{\textbf{MLT}$_A$} \\ \hline
RCSLS & 37.67 & 77.59 & \textbf{87.05} & 22.23 & \textbf{59.12} & 57.75 & 35.12 & 68.63 & \textbf{81.44} & 8.72 & 29.44 & \textbf{30.42} \\ \hline
XLM (MLM) & 60.8 & 75.11 & \textbf{83.95} & 38.55 & 63.29 & \textbf{66.11} & 37.59 & 46.34 & \textbf{65.31} & 8.12 & 19.03 & \textbf{20.43} \\
\quad+ Transformer & 62.33 & 82.83 & \textbf{85.63} & 41.67 & 66.53 & \textbf{67.95} & 40.31 & 57.27 & \textbf{68.55} & 11.45 & 26.02 & \textbf{27.45} \\ \hline
XLM (TLM+MLM) & 62.48 & 81.34 & \textbf{84.91} & 42.27 & 65.71 & \textbf{66.48} & 31.62 & 50.34 & \textbf{65.25} & 7.91 & 19.22 & \textbf{19.88} \\
\quad+ Transformer & 65.32 & 83.79 & \textbf{87.48} & 44.39 & 66.03 & \textbf{68.55} & 37.53 & 68.62 & \textbf{72.59} & 12.84 & 26.56 & \textbf{27.98} \\ \hline
Multi. BERT & 73.73 & 77.51 & \textbf{86.54} & 51.73 & \textbf{74.51} & 74.43 & 28.15 & 52.25 & \textbf{70.57} & 10.62 & 24.41 & \textbf{28.47} \\
\quad+ Transformer & 74.15 & 82.9 & \textbf{87.88} & 54.28 & \textbf{74.88} & 73.89 & 26.54 & 53.84 & \textbf{73.46} & 11.34 & 26.05 & \textbf{27.12} \\ \hline  \hline
\textit{Zero-shot SLU$^\dagger$} & \multicolumn{3}{c|}{\textit{46.64}} & \multicolumn{3}{c|}{\textit{15.41}} & \multicolumn{3}{c|}{\textit{35.64}} & \multicolumn{3}{c}{\textit{12.11}} \\
\textit{Multi. CoVe} & \multicolumn{3}{c|}{\textit{53.34}} & \multicolumn{3}{c|}{\textit{22.50}} & \multicolumn{3}{c|}{\textit{66.35}} & \multicolumn{3}{c}{\textit{32.52}} \\
\textit{Multi. CoVe w/ auto} & \multicolumn{3}{c|}{\textit{53.89}} & \multicolumn{3}{c|}{\textit{19.25}} & \multicolumn{3}{c|}{\textit{70.70}} & \multicolumn{3}{c}{\textit{35.62}} \\ \hline
\textit{Translate Train} & \multicolumn{3}{c|}{\textit{85.39}} & \multicolumn{3}{c|}{\textit{72.87}} & \multicolumn{3}{c|}{\textit{95.85}} & \multicolumn{3}{c}{\textit{55.43}} \\ \hline
\end{tabular}
}
\caption{Results on multilingual NLU dataset~\cite{schuster2019cross}, and the number of word pairs on both MLT$_H$ and MLT$_A$ is 20. $^\dagger$ We implemented the model~\cite{upadhyay2018almost} and tested it on the same dataset.}
\label{table:nlu}
\end{table*}

\subsection{Experimental Setup}
We explore two training settings: (1) without Mixed-language Training (BASE), and (2) Mixed-language Training (MLT). The former trains models only using English data, and then we directly transfer to the target language by leveraging the same cross-lingual word embeddings as our model. The latter utilizes code-switching sentences as the train data. We evaluate our model with cross-lingual embeddings: MUSE~\cite{conneau2017word}, RCSLS~\cite{joulin2018loss}, XLM~\cite{lample2019cross}, and Multilingual BERT (Multi. BERT)~\cite{devlin2019bert}. 

We describe our baselines for the \textit{dialogue state tracking} task in the following:
\subsubsection{Ontology-based Word Selection (MLT$_{O}$)} We use dialogue ontology word pairs for mixed-language training since ontology words are all task-related and essential for the DST task.
\subsubsection{XL-NBT} \citet{chen2018xl} proposed a teacher-student framework for cross-lingual neural belief tracking (i.e., dialogue state tracking) by leveraging a bilingual corpus or bilingual dictionary. The model learns to generate close representations for semantically similar sentences across languages.
\subsubsection{Ontology Matching} \citet{chen2018xl} directly used exact string matching for the user utterance according to the ontology words to discover the slot value for each slot type.
\subsubsection{Translate Train} \citet{chen2018xl} used an external bilingual corpus to train a machine translation system, which translates English dialogue training data into target languages (German and Italian) as ``annotated'' data to supervise the training of DST systems in target languages.
\subsubsection{Supervised Training} We assume the existence of annotated data for the target languages dialogues state tracking. It indicates the upper bound of the DST model.

\noindent\paragraph{}We describe our baselines for the \textit{natural language understanding} task in the following:

\subsubsection{Human-based Word Selection (MLT$_H$)} Due to the absence of ontology in the NLU task, we crowd-source the top-20 task-related source words in the English training set.
\subsubsection{Zero-shot SLU} \citet{upadhyay2018almost} used cross-lingual word embeddings~\cite{bojanowski2017enriching} to conduct zero-shot transfer learning in the NLU task.
\subsubsection{Multi. CoVe} \citet{schuster2019cross} used Multilingual CoVe~\cite{yu2018multilingual} to encode phrases with similar meanings into similar vector spaces across languages.
\subsubsection{Multi. CoVe w/ auto.} Based on Multilingual CoVe, \citet{schuster2019cross} added an autoencoder objective to produce more general representations for semantically similar sentences across languages.
\subsubsection{Translate Train} \citet{schuster2019cross} trained a machine translation system using a bilingual corpus, and then translated English NLU data into the target languages (Spanish and Thai) for supervised training.

\subsection{Evaluation Metrics}
\subsubsection{Dialogue State Tracking}
We use \textit{joint goal accuracy} and \textit{slot accuracy} to evaluate the model performance on goal-tracking slots. The joint goal accuracy compares the predicted dialogue states to the ground truth at each dialogue turn, and the prediction is correct if and only if the predicted values for all slots exactly match the ground truth values. While the slot accuracy individually compares each slot-value pair to its ground truth. We use \textit{request accuracy} to evaluate the model performance on ``request" slot. Similar to joint goal accuracy, the prediction is correct if and only if all the user requests for information are correctly identified.

\subsubsection{Natural Language Understanding}
We use \textit{accuracy} and \textit{BIO-based f1-score} to evaluate the performance of intent prediction and slot filling, respectively.

\begin{figure*}[!ht]
\centering
\begin{subfigure}{.24\textwidth}
    \centering
    \includegraphics[scale=0.32]{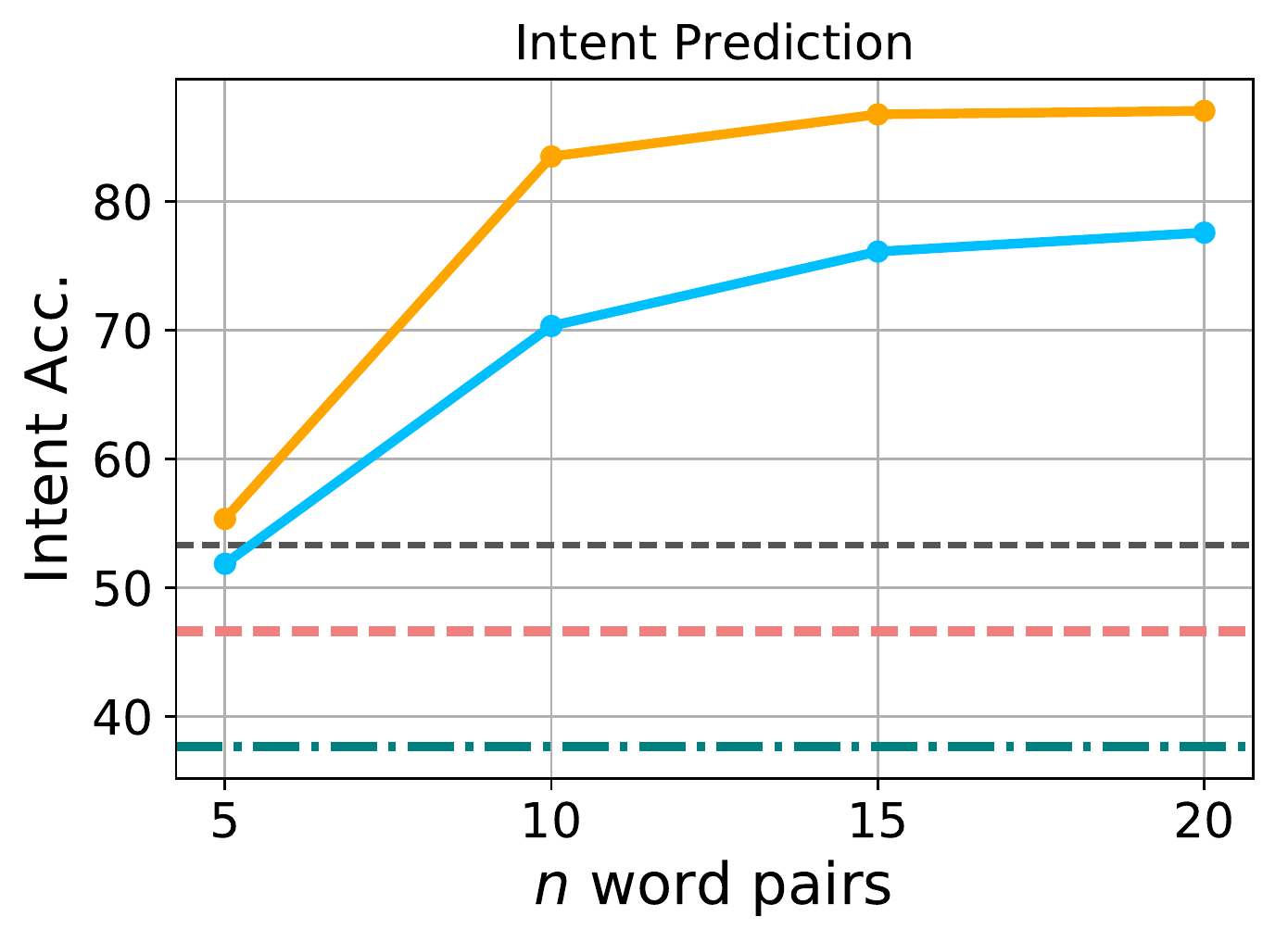}
    \caption{}
    \label{fig:intent-es}
\end{subfigure}
\begin{subfigure}{.24\textwidth}
    \centering
    \includegraphics[scale=0.32]{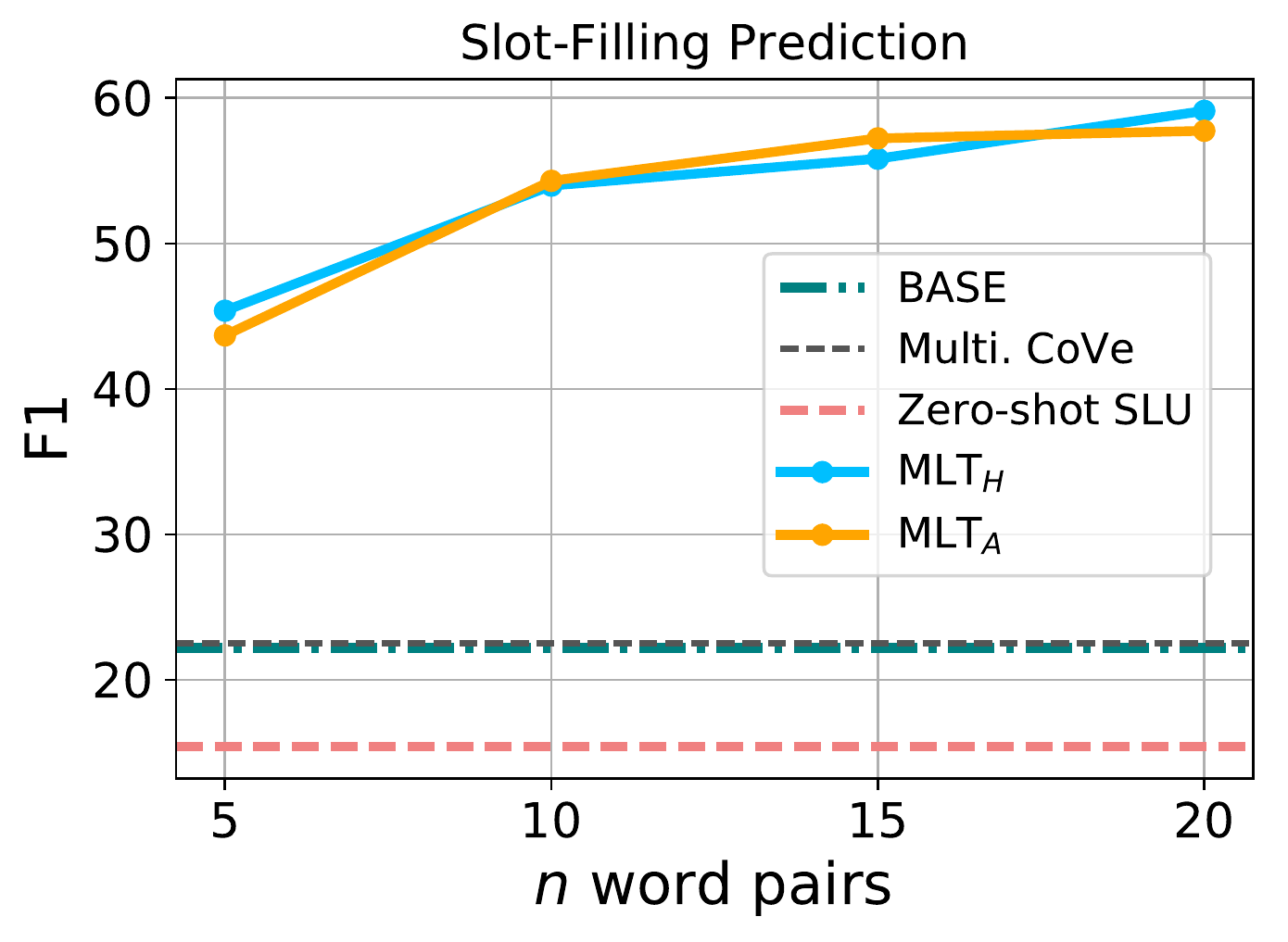}
    \caption{}
    \label{fig:slot-es}
\end{subfigure}
\begin{subfigure}{.24\textwidth}
    \centering
    \includegraphics[scale=0.32]{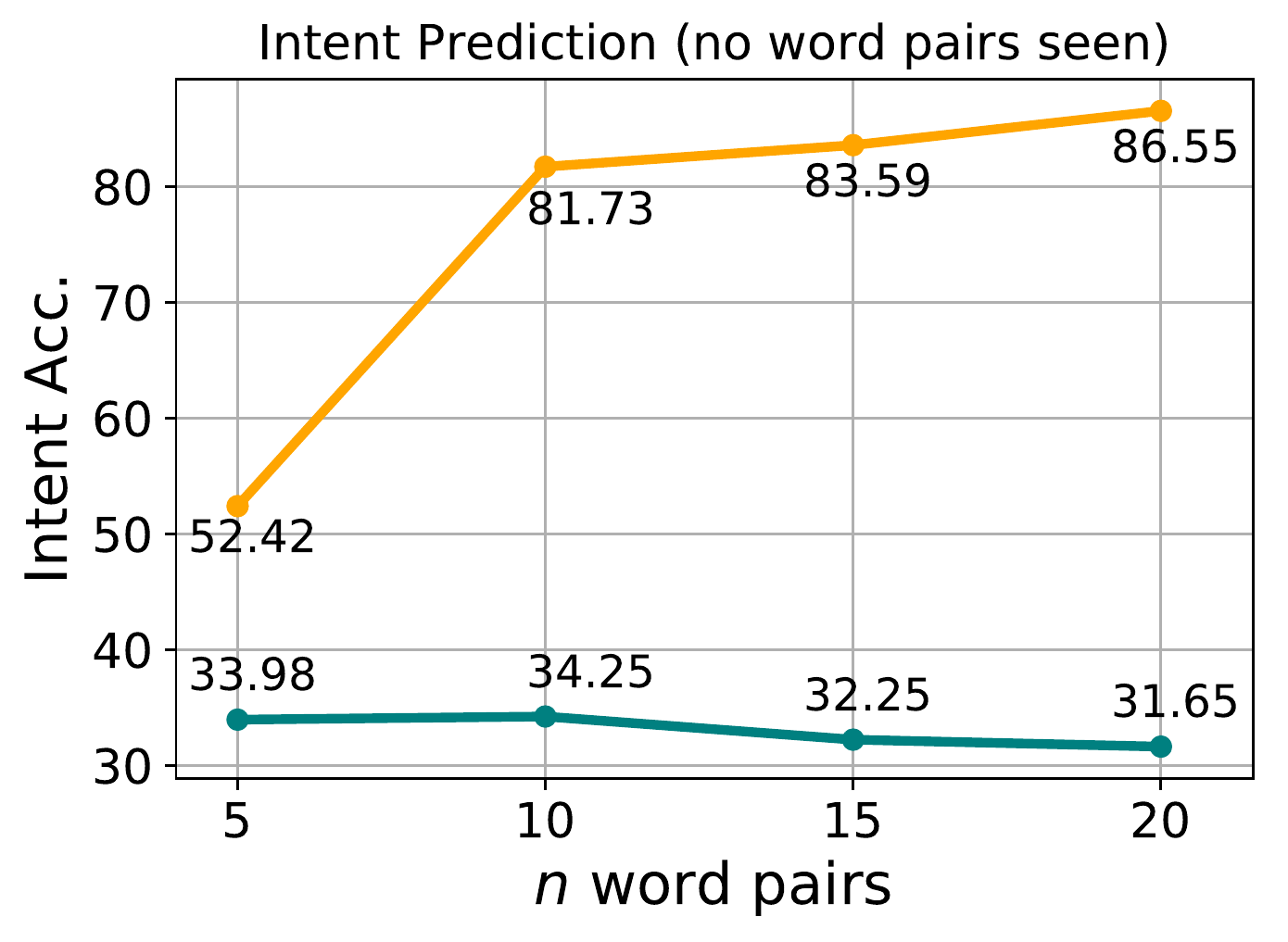}
    \caption{}
    \label{fig:intent-es-unseen}
\end{subfigure}
\begin{subfigure}{.24\textwidth}
    \centering
    \includegraphics[scale=0.32]{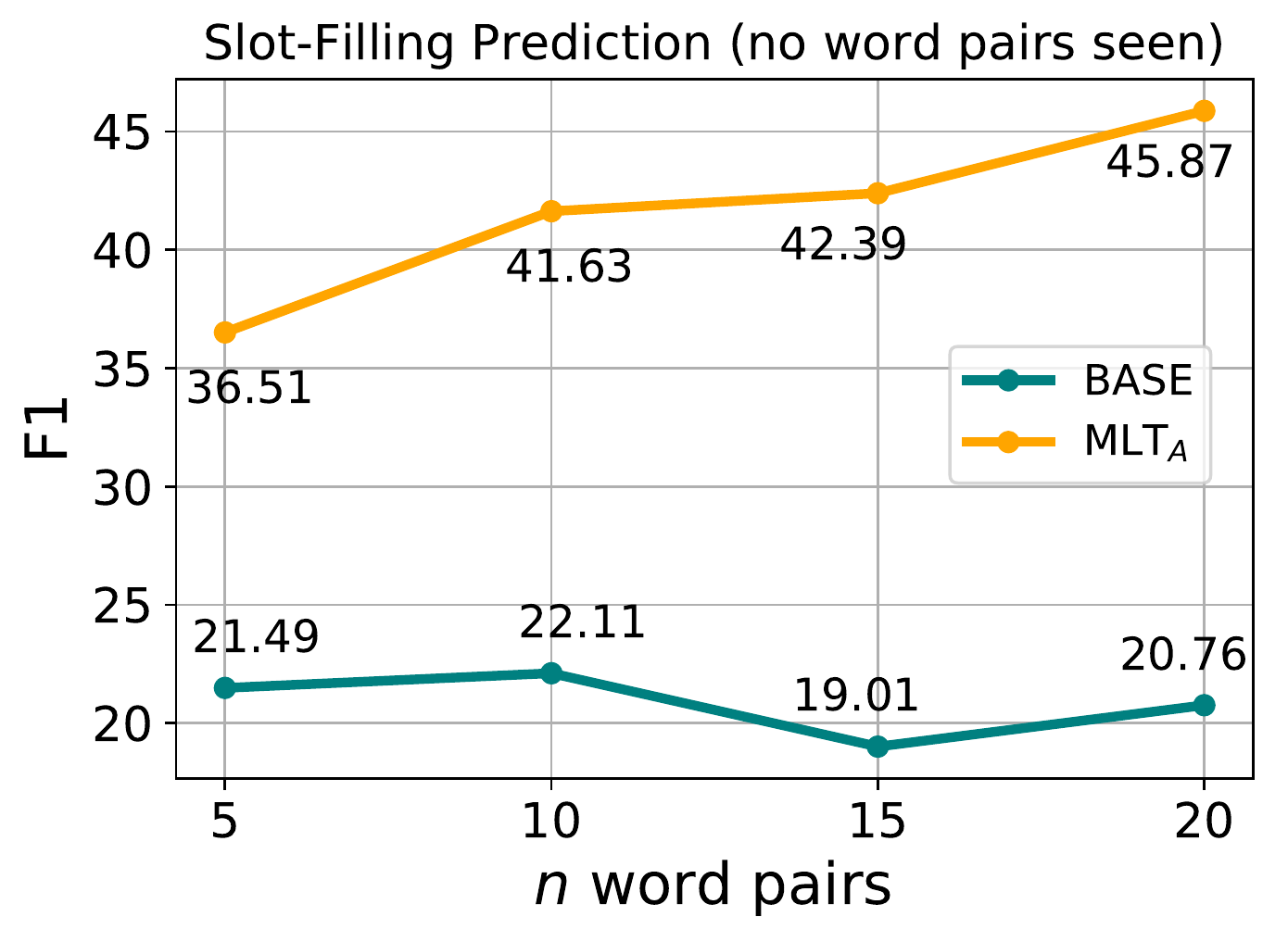}
    \caption{}
    \label{fig:slot-es-unseen}
\end{subfigure}
\caption{The dynamics of the NLU task: intent and slot-filling results with different numbers of word pairs on Spanish test data using RCSLS. The words are decided according to the frequency in the source language (English) training set. We evaluate on all test data for \textbf{(a)} and \textbf{(b)}. For \textbf{(c)} and \textbf{(d)}, we only evaluate on filtered test data that do not contain any word pairs.}
\label{fig:dynamics}
\end{figure*}

\begin{figure*}[!ht]
\begin{subfigure}{.94\textwidth}
    \centering
    \includegraphics[scale=0.21]{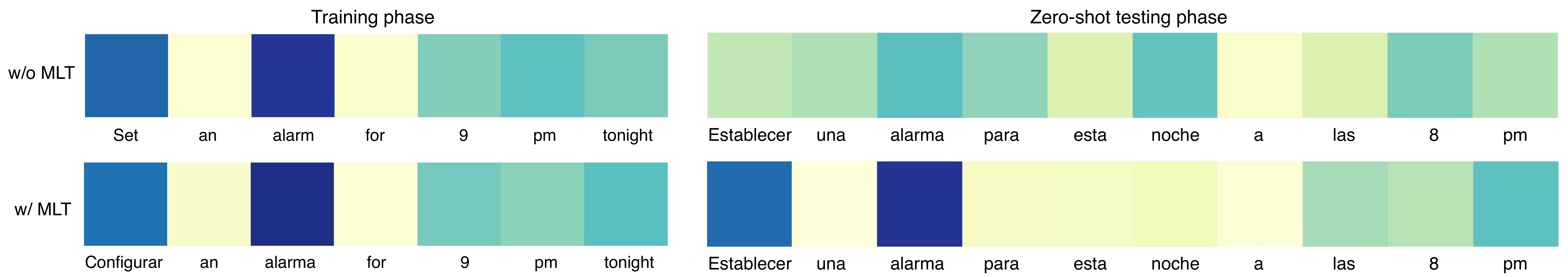}
\end{subfigure}
\begin{subfigure}{.05\textwidth}
    \centering
    \includegraphics[scale=0.3, height=98pt]{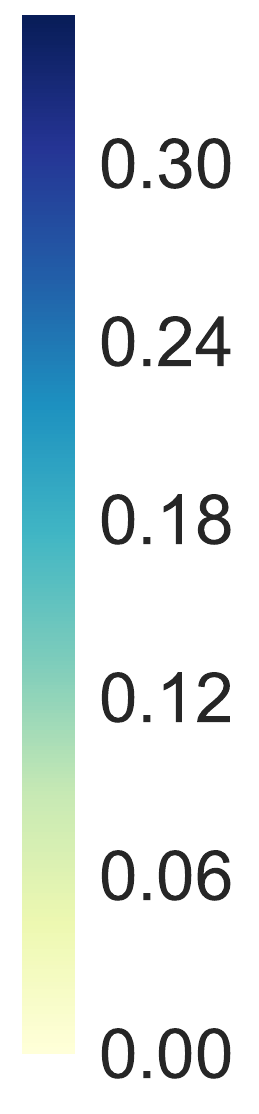}
    \label{fig:bar}
\end{subfigure}
\caption{Attentions on words in both training and testing phases. A darker color shows a higher attention score and importance.}
\label{fig:attention}
\end{figure*}

\section{Results \& Discussion}
\subsection{Quantitative Analysis}
% MLT outperforms all baselines -> showing that our model is able to utilize task-related words from different languages
The DST and NLU results are shown in Table \ref{table:dst} and \ref{table:nlu}. In most cases, our models using MLT significantly outperform the existing state-of-the-art zero-shot baselines, and we achieve a comparable result to the \textit{Multi. CoVe w/ auto} on Thai. Notably, our models achieve impressive performance since we only use a few word pairs and many fewer bilingual resources than sophisticated models such as \textit{Multi. Cove} or \textit{Bilingual Corpus}.
 
We observe that \textit{ontology matching} is an intuitive method to attempt zero-shot in low-resource languages. However, this method is ineffective because it does not seem able to detect synonyms or paraphrases. Applying ontology pairs into the MLT models copes with this problem and outperforms the BASE models with vast improvements.
Interestingly, MLT$_A$ consistently outperforms MLT$_O$ because the attention-based selection mechanism 
is not only capturing important ontology keywords but also keywords which are not listed in the ontology (i.e., synonyms or paraphrases to the ontology words). For example, word ``moderate" is interchangeable with ``fair" when users describe the food price during the conversation, which is not listed in the ontology.
Since we do not have an ontology in the NLU task, we compare our results with human crowd-sourcing-based word selection (MLT$_H$). Results show that MLT$_A$ significantly outperforms human word pairs selection MLT$_H$ in the intent detection, which further proves the high quality of words selected by the attention layer.

Due to the imperfect alignment of cross-lingual word embeddings, our BASE models with MUSE or RCSLS still suffer from low performance in the zero-shot adaptation. 
Although we replace these cross-lingual word embeddings with large pre-trained language models such as XLM and Multi. BERT, the performance is not consistently better. This is because the quality of alignment degrades when we combine subword-based embeddings into word-level representations. The performance of the XLM-based models and Multi. BERT-based models are improved remarkably by applying MLT. Surprisingly, MLT-based models with RCSLS surpass XLM and Multi. BERT by a substantial margin on the Thai language. We find that the length of Thai subword sequences is approximately twice as long as other languages. Hence, the quality of subword-to-word alignments degrades severely.

\subsection{Performance vs. Number of Word Pairs}
Figure \ref{fig:intent-es} and \ref{fig:slot-es} compare the performance of intent and slot-filing predictions on Spanish data with respect to the number of word pairs, and investigates the gap between \textit{human crowd-sourcing-based word selection} (MLT$_H$) and \textit{attention-based word selection} (MLT$_A$). Interestingly, with only five word pairs, MLT$_A$ achieves notable gains of 17.69\% and 21.45\% in intent prediction and slot filling performance, respectively, compared to the BASE model. Compared with human word pairs selection MLT$_H$, in the intent prediction, MLT$_A$ beats the performance of human-based word selection, and in slot-filling prediction, the result is on par with the MLT$_H$.

\subsection{Model Transferability}
In Figure \ref{fig:intent-es-unseen} and \ref{fig:slot-es-unseen}, we show the transferability of MLT$_A$ on the target language data that does not have any target keywords selected from the word pair list. Our model with MLT$_A$ is still able to achieve impressive gains on both intent and slot-filling performance on these data. The results emphasize that the MLT-based model not only memorizes target word replacements, but captures the generic semantics of words and learns to generalize to other words that have a similar vector space, for example, the synonyms ``configurer'' and ``establecer'' (both mean ``set'' in English) or word from the same domain, like ``Domingo'' (Sunday) and ``Lunes'' (Monday).

To further support our claims, we extract the attention scores from the attention layer and elaborate on the findings. Figure \ref{fig:attention} displays that, in the training phase, our model puts attentions on parallel task-related words in both the source and target languages, such as ``Set" and ``alarm" in English, and ``Configurar" and ``alarma" in Spanish. In the zero-shot test phase, our attention layer in the MLT-based models puts an attention on identical or synonym words because they have the same or similar vector representations, respectively, but without MLT, our attention layer fails to do so.
Interestingly, we can see clearly in Figure \ref{fig:attention} that word `Establecer" is as equally important as ``Configurer", although ``Establecer" is not found in the code-switching sentence.

\section{Conclusion}
We propose attention-informed mixed-language training (MLT), a novel zero-shot adaptation method for cross-lingual task-oriented dialogue systems using code-switching sentences. Our approach utilizes very few task-related parallel word pairs based on the attention layer and has a better generalization to words that have similar semantics in the target language. The visualization of the attention layer confirms this. Experimental results show that MLT-based models outperform existing zero-shot adaptation approaches in dialogue state tracking and natural language understanding with many fewer resources.

% \bigskip
% \noindent Thank you for reading these instructions carefully. We look forward to receiving your electronic files!
% \clearpage

\bibliographystyle{aaai}
\bibliography{aaai2020}

\end{document}